\documentclass[a4paper, 10pt, conference]{ieeeconf}      

\IEEEoverridecommandlockouts                              

\pdfoutput=1

\usepackage{graphics} 
\usepackage{epsfig} 
\usepackage{mathptmx} 
\usepackage{times} 
\usepackage{amsmath} 
\usepackage{amssymb}  
\usepackage{multirow}
\usepackage{graphicx}
\usepackage[table,xcdraw]{xcolor}
\usepackage{xcolor}
\usepackage{float}
\usepackage{cite}

\title{\LARGE \bf
You Sense Only Once Beneath: Ultra-Light Real-Time Underwater Object Detection
}

\author{Jun Dong$^{1}$, Wenli Wu$^{1}$, Jintao Cheng$^{2}$ and Xiaoyu Tang$^{3}$
\thanks{$^{1}$School of Data Science and Engineering, and Xingzhi College, South China Normal University, Shanwei}%
\thanks{$^{2}$School of Physics, South China Normal University, Guangzhou}%
\thanks{$^{3}$School of Electronics and Information Engineering, and Xingzhi College, South China Normal University, Shanwei}%
}

\begin{document}

\maketitle
\thispagestyle{empty}
\pagestyle{empty}

\begin{abstract}

Despite the remarkable achievements in object detection, the model's accuracy and efficiency still require further improvement under challenging underwater conditions, such as low image quality and limited computational resources. To address this, we propose an Ultra-Light Real-Time Underwater Object Detection framework, You Sense Only Once Beneath (YSOOB). Specifically, we utilize a Multi-Spectrum Wavelet Encoder (MSWE) to perform frequency-domain encoding on the input image, minimizing the semantic loss caused by underwater optical color distortion. Furthermore, we revisit the unique characteristics of even-sized and transposed convolutions, allowing the model to dynamically select and enhance key information during the resampling process, thereby improving its generalization ability. Finally, we eliminate model redundancy through a simple yet effective channel compression and reconstructed large kernel convolution (RLKC) to achieve model lightweight. As a result, forms a high-performance underwater object detector YSOOB with only 1.2 million parameters. Extensive experimental results demonstrate that, with the fewest parameters, YSOOB achieves mAP50 of 83.1\% and 82.9\% on the URPC2020 and DUO datasets, respectively, comparable to the current SOTA detectors. The inference speed reaches 781.3 FPS and 57.8 FPS on the T4 GPU (TensorRT FP16) and the edge computing device Jetson Xavier NX (TensorRT FP16), surpassing YOLOv12-N by 28.1\% and 22.5\%, respectively.

\end{abstract}

\section{INTRODUCTION}

\begin{figure}[]
\centering
\includegraphics[width=3.5in]{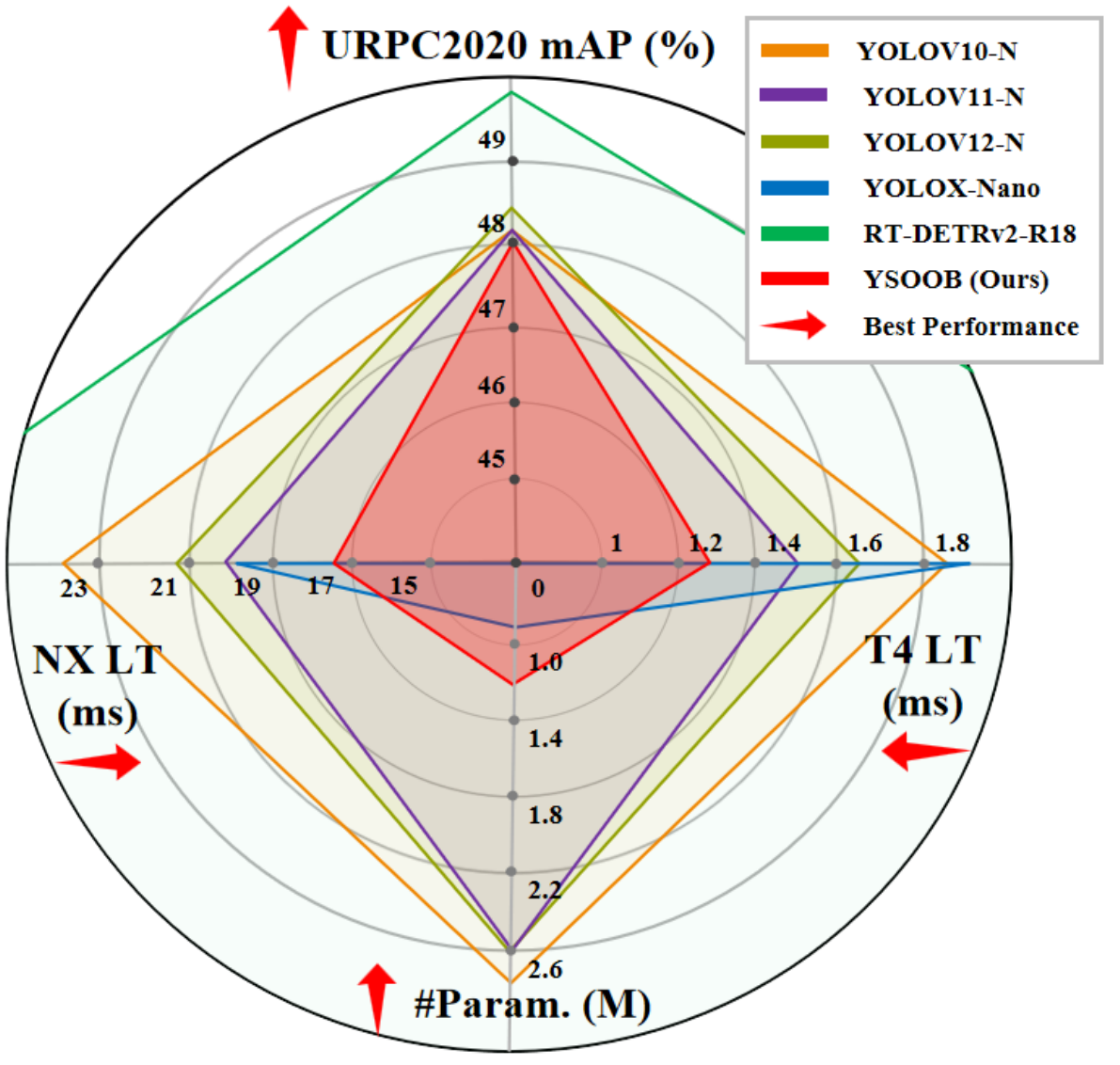}
\caption{Comparison of accuracy (top), parameters (bottom), and inference latency (TensorRT FP16) on Jetson Xavier NX (left) and T4 GPU (right) against other popular SOTA methods. The red arrows in the radar chart indicate the direction of optimal extension for the model.}
\label{fig_1}
\end{figure}

\begin{figure*}[!htbp]
\centering
\includegraphics[width=6.2in]{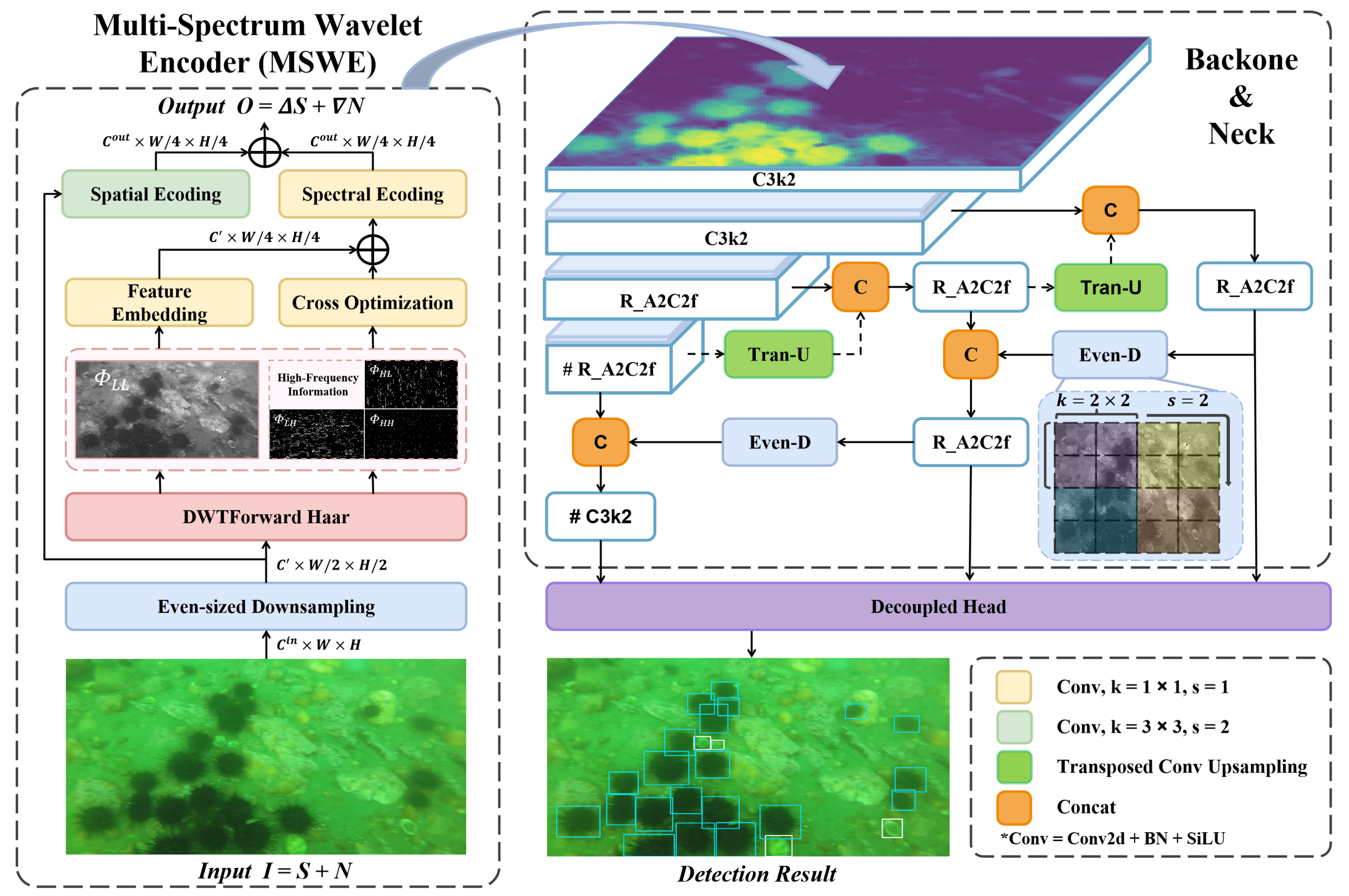}
\caption{Overall architecture of YSOOB. '\#' and 'R\_' represent the channel compression operation and RLKC. For the input image I, S represents the target signal, and N represents the additive noise.}
\label{fig_1}
\end{figure*}

Deep learning has made remarkable advancements in fields such as autonomous driving\cite{stereo, ViPOcc, mfmos}, with real-time object detection technologies, exemplified by YOLOs and DETRs, gaining widespread adoption \cite{1, leps}. However, due to factors such as light attenuation, color distortion, and difficulty distinguishing targets from coral reefs, mud, and other underwater structures, the development of high-performance real-time underwater object detection (UOD) has been relatively slow \cite{3}. Moreover, embedded devices' limited storage and computational capabilities make it challenging to deploy models similar to Vision Transformers (ViT)\cite{edgeant}. Achieving an effective balance between accuracy and efficiency in underwater object detection remains a critical challenge.

A common strategy for improving UOD performance involves underwater image enhancement (UIE) for image preprocessing. However, the primary goal of UIE is to enhance visual quality, and this approach not only increases computational overhead and latency but may also introduce artifacts that, in turn, degrade detection performance \cite{5}. Some researchers have attempted to combine image enhancement with object detection end-to-end, building multitask networks to alleviate the computational burden \cite{6,7}. However, these methods still face challenges, such as increased model complexity and a heavy reliance on synthetic data for supervised learning, leading to insufficient generalization ability in real-world applications.

In this work, we propose an Ultra-Light Real-Time Underwater Object Detection method, YSOOB, which aims to eliminate the reliance on UIE through image frequency-domain signal processing, achieving greater robustness in real-world applications. YSOOB extends the attention-centered YOLO framework and further achieves model lightweight through resampling, large kernel convolution reconstruction, and channel compression, striking an excellent balance between real-time inference speed and detection accuracy. Fig. 1 provides a detailed comparison between YSOOB and other models. The contributions are summarized as follows.\begin{enumerate}

\item{A multi-spectrum wavelet encoder (MSWE) is introduced to efficiently extract features and reconstruct degraded underwater images by leveraging frequency-domain feature transformations, thus eliminating the dependency on UIE processing. Additionally, the dynamic perceptual properties of even-sized and transposed convolutions are re-examined as viable replacements for traditional upsampling and downsampling operations, effectively reducing the loss of edge features while cutting down on model parameters. To further streamline the model, a simple yet effective channel compression and reconstructed large kernel convolution (RLKC) to remove redundancy, achieving the final lightweight model design while maintaining sensitivity to small targets.}

\item{Experimental results show that YSOOB, with only 1.2 million parameters, achieves accuracies of 83.1\% and 82.9\% on the URPC2020 and DUO datasets. It performs comparably to the baseline model YOLOv12n while reducing the number of parameters and FLOPs by 53.85\% and 25.40\%. Furthermore, YSOOB achieves impressive real-time inference speeds of 781.3 FPS on the T4 GPU and 57.8 FPS on the edge computing device Jetson Xavier NX.}
\end{enumerate}

\section{RELATED WORK}

\subsection{Lightweight Underwater Object Detection}

Several approaches have achieved remarkable accuracy in underwater object detection within complex environments \cite{8,9}; however, they often overlook model portability and do not enable rapid detection. Li et al. \cite{10} optimized the YOLOv5 backbone through network pruning and distillation techniques and employed specialized network search algorithms to develop an efficient model that improved detection speed by 12 times. Ouyang et al. \cite{12} developed the lightweight feature extraction network Mobile-bone, resulting in a final DU-MobileYOLO model with only 4.7M parameters. Cheng et al. \cite{13}, building upon YOLOv11n, introduced a multi-sampling point mechanism to enhance efficiency, with the final model having only 2.58M parameters. Based on previous studies, this paper proposes an Ultra-Light Real-Time YOLO detector that balances detection performance with model efficiency.

\subsection{Image Frequency-Domain Analysis}

With the advancement of deep learning, frequency domain transformations \cite{14} and optical flow descriptions \cite{dcpi} have been increasingly applied in visual perception. Xu et al. \cite{15} proposed a Haar Wavelet Downsampling module, which can be seamlessly integrated into CNNs. In response to the degradation issues of underwater optical images, frequency domain transformations have gradually emerged as an alternative for image enhancement. Zhang et al. \cite{16} optimized feature maps through techniques such as frequency domain weighting. Zhao et al. \cite{17} thoroughly explored the frequency domain features of the original image pixels, performing preliminary enhancement of the underwater image frequency information in the wavelet domain.



\section{METHOD} 

This section provides a detailed overview of the proposed method YSOOB. We use the SOTA single-stage object detector YOLOv12-N \cite{v12} as the baseline model to balance real-time performance and accuracy. The overall framework of YSOOB is shown in Fig. 2. The Multi-Spectrum Wavelet Encoder (MSWE) first processes the input image for frequency domain feature extraction and encoding reconstruction without relying on image enhancement techniques. Next, we revisited the dynamic perceptual characteristics of even-sized and transpose convolutions, replacing all downsampling and upsampling operations in YOLOs framework. This significantly optimizes the model parameters and reduces the loss of object edge features. Finally, a simple yet effective channel compression and reconstructed large kernel convolution (RLKC) are applied to eliminate model redundancy, achieving the final Ultra-Light design.

\subsection{Multi-Spectrum Wavelet Encoder}

In signal communication, spectrum analysis can effectively eliminate noise interference. Compared to UIE methods, frequency domain analysis of images has significant advantages in real-time underwater target detection tasks. Inspired by \cite{15}, we propose the Multi-Spectrum Wavelet Encoder (MSWE), which efficiently performs space-frequency joint embedding and encoding reconstruction of the original input underwater degraded image. To reduce the image resolution during frequency domain transformation while maintaining the translational invariance of the original image spatial features as much as possible, for a given input image $I \in \mathbb{R}^{H \times W \times C}$, we directly apply a $2\times2$ convolution with a stride of 2 for downsampling to obtain feature $X_0 \in \mathbb{R}^{H / 2 \times W / 2 \times C^{\prime}}$, where $C^{\prime}$ is the intermediate hidden channel number. Subsequently, we perform a 2D discrete wavelet transform using the Haar wavelet basis function to obtain low-frequency components $Y_{\mathrm{L}}$ and three high-frequency components $Y_{\mathrm{HL}}$, $Y_{\mathrm{LH}}$, and $Y_{\mathrm{HH}}$, which correspond to the detail features in the horizontal, vertical, and diagonal directions, respectively. In response to scattering noise interference, we cross-optimize high-frequency feature representations:

\begin{equation}
\tilde{Y}_H=\mathcal{F}_{\text {compress }}\left(\left[Y_{\mathrm{HL}} \oplus Y_{\mathrm{LH}} \oplus Y_{\mathrm{HH}}\right]\right)
\end{equation}
where $\oplus$ denotes channel concatenation, and $\mathcal{F}_{\text {compress }}$ represents the convolution operation. The enhanced high-frequency features are combined with low-frequency signals to perform dual-path spectrum interaction, resulting in the reconstructed spectral feature $\mathcal{P}_{\text {freq }}$. After channel mapping, the spatial structural position encoding information is aligned to produce $\mathcal{P}_{\text {res }}$. Finally, the feature fusion output $Z$ is obtained by constructing a path dominated by low-frequency components and a residual path.

\begin{equation}
\mathcal{P}_{\text {freq }}, \mathcal{P}_{\text {res }}=\mathcal{C}_1\left(Y_L+\tilde{Y}_H\right), \mathcal{C}_3^{\downarrow}\left(\mathrm{X}_0\right)
\end{equation}

\begin{equation}
Z=\mathcal{P}_{\text {low }} \oplus \mathcal{P}_{\text {res }}
\end{equation}
where $\mathcal{C}_3^{\downarrow}$ denotes the convolution and secondary downsampling operations in this context, and $\mathcal{C}_1$ achieves channel dimension mapping. After high-frequency noise suppression, the final output retains both the original spatial structural feature information and the frequency spectral features. For underwater degraded images with noise input $\mathrm{I}=\mathrm{S}+\mathrm{N}$ after wavelet decomposition:

\begin{equation}
\left\|Y_H\right\|_F \propto\|\nabla S+\nabla N\|_2
\end{equation}

Through the feature cross-adjustment of high-frequency channels, the model can adaptively suppress high-frequency noise components that satisfy $\|\nabla N\|_2>\|\nabla S\|_2$. MSWE effectively separates high-frequency water scattering noise from low-frequency target signal components using a wavelet transform, thereby improving the model's ability to resist interference from aquatic media and enhancing its robustness in complex underwater environments.

\subsection{Resampling}
\subsubsection{Upsampling}

In YOLO's framework, the upsampling method typically relies on nearest-neighbour interpolation for linearly resizing the deep feature maps, which fails to adapt to the complex variations in image content. As a standard operation in Generative Adversarial Networks (GANs) \cite{19}, transpose convolution uses learnable dynamic kernels to maximally and adaptively recover high-frequency details such as edges, textures, and delicate object contours during the deconvolution process. We replace all upsampling operations in the original framework with $2\times2$ transpose convolutions. Parametric upsampling allows the model to learn the fusion of the target's deep and shallow positional information, effectively addressing the severe high-frequency degradation caused by light scattering and suspended particles in underwater images, thereby enhancing the model's ability to perceive small underwater targets. Compared to traditional methods, the increase in model parameters and complexity is negligible.

\begin{table*}[ht]
\centering
\caption{COMPARISON WITH OTHER DETECTORS ON URPC2020 VAL SET}
\label{tab:combined-table}
\renewcommand{\arraystretch}{1.3}
\resizebox{2.0\columnwidth}{!}{
\begin{tabular}{ccccccccccccccc}
\hline
\multirow{2}{*}{\textbf{Model}} & 
  \multirow{2}{*}{\textbf{\begin{tabular}[c]{@{}c@{}}mAP50 \\ (\%)\end{tabular}}} & 
  \multirow{2}{*}{\textbf{\begin{tabular}[c]{@{}c@{}}mAP \\ (\%)\end{tabular}}} & 
  \multicolumn{4}{c}{\textbf{mAP (\%)}} & \multirow{2}{*}{\textbf{\begin{tabular}[c]{@{}c@{}}\#Param. \\ (M)\end{tabular}}} & 
  \multirow{2}{*}{\textbf{\begin{tabular}[c]{@{}c@{}}FLOPs \\ (G)\end{tabular}}} & 
  \multirow{2}{*}{\textbf{\begin{tabular}[c]{@{}c@{}}$\text{Latency}_{\text{T4}}^{\text{TRT}}$ \\ (ms)\end{tabular}}} & 
  \multirow{2}{*}{\textbf{\begin{tabular}[c]{@{}c@{}}$\text{Latency}_{\text{NX}}^{\text{TRT}}$ \\ (ms)\end{tabular}}} \\ \cline{4-7}
                                         &      &      & \textbf{Holothurian} & \textbf{Echinus} & \textbf{Starfish} & \textbf{Scallop} \\ \hline
YOLOv12-N \cite{v12} (Baseline)   & 83.2 & \underline{48.3} & \underline{38.8}  & 52.7 & 52.4 & 49.3 & 2.6  & 6.3  & 1.64 & 21.2 \\
YOLOv8-N \cite{v8v11}      & \underline{83.5} & 47.9 & 38.0  & 53.4 & 51.6 & 48.6 & 3.0  & 8.2  & 1.77 & 21.4 \\
YOLOv10-N \cite{v10} & 83.4 & 48.1 & 36.8  & 52.9 & 52.2 & 50.5 & 2.7  & 8.4  & 1.84 & 23.7 \\
YOLOv11-N \cite{v8v11} & 83.0 & 48.1 & 38.4  & \underline{53.2} & 51.4 & \underline{49.8} & 2.6  & 6.3  & \underline{1.50} & 20.9 \\
YOLOX-Nano \cite{yolox}    & 48.8 & 21.3 & 16.7  & 24.1 & 23.4 & 21.0 & \textbf{0.9}  & \textbf{1.1}  & 1.89 & \underline{19.7} \\
RT-DETRv2-R18 \cite{rtdetrv2} & \textbf{85.4} & \textbf{49.7} & \textbf{39.7}  & \textbf{54.4} & \textbf{53.8} & \textbf{50.9} & 20.0 & 60.0 & 4.58 & 42.5 \\
Faster-RCNN \cite{faster}   & 75.3 & 40.2 & 33.7  & 43.8 & 42.7 & 40.6 & 41.3 & 90.9 & -    & -    \\
\rowcolor{gray!20} \textbf{YSOOB (Ours)}  & 83.1 & 48.0 & 38.7  & 52.5 & \underline{52.7} & 47.9 & \underline{1.2}   & \underline{4.7}   & \textbf{1.28}   & \textbf{17.3}   \\ \hline
\end{tabular}%
}
\end{table*}

\subsubsection{Downsampling}
In YOLOs, downsampling is typically performed using a $3\times3$ convolution with a stride of 2. However, we believe that a dedicated feature extraction module should handle the task of capturing continuous spatial representations. Given that underwater targets are often small, blurred, and susceptible to noise, excessive feature map compression can lead to the loss of edge and detail information, resulting in overfitting. Therefore, we replace all downsampling operations in YOLOs with $2\times2$ convolutions with a stride of 2, which use only 44\% of the parameters of $3\times3$ convolution. The even convolution operation enables a more refined focus on local features, helping to reduce spatial information loss, improve feature map alignment, and enhance the model's generalization capability.

\subsection{Model Parameter Optimization}
\subsubsection{Reconstructed Large-Kernel Convolution}
Underwater images suffer from spatial blur and colour distortion between channels caused by light scattering and suspended particles. To address the underfitting caused by insufficient model parameters, we enhance the backbone by using large kernel depthwise separable convolution layers to cover a broader degraded region, thereby strengthening the multi-scale contextual feature reconstruction ability between spatial and channel dimensions. For input features:

\begin{equation}
\mathrm{F}_{\mathrm{dw}}(\mathrm{x}, \mathrm{y}, \mathrm{c})=\sum_{\mathrm{i}, \mathrm{j}=-3}^3 \mathrm{~W}_{\mathrm{dw}}(\mathrm{c}, \mathrm{i}+3, \mathrm{j}+3) \cdot \mathrm{X}(\mathrm{x}+\mathrm{i}, \mathrm{y}+\mathrm{j}, \mathrm{c})
\end{equation}

\begin{equation}
\mathrm{F}_{\mathrm{pw}}\left(\mathrm{c}^{\prime}\right)=\sum_{\mathrm{c}=1}^{\mathrm{c}_{\mathrm{in}}} \mathrm{~W}_{\mathrm{pw}}\left(\mathrm{c}^{\prime}, \mathrm{c}\right) \cdot \mathrm{F}_{\mathrm{dw}}(\mathrm{c})
\end{equation}
where $w_1$ and $w_2$ represent the depthwise and pointwise convolution kernels, respectively, c is the channel index, while i and j denote the spatial offset positions of the convolution kernel. The reconstructed convolution embedding provides prior spatial constraints that complement the area-attention mechanism and helps mitigate the attention weight divergence caused by water blur.

\subsubsection{Channel Compression}
When the total number of model parameters falls below a certain threshold, the performance gains from channel expansion exhibit a significant diminishing marginal effect, with miniature models having up to 40-60\% channel redundancy \cite{mobilenetv3,channel}. Considering that the YOLOv12 backbone architecture, which incorporates attention mechanisms, has sufficient capacity to capture the most valuable information, we simply reduce the width of the N-sized model backbone's final layer and the neck's last layer to half of their original size. Removing redundant parameters allows the network to focus more on core information, improving training and inference efficiency.

\section{EXPERIMENT}

This section is divided into three parts: experimental setup, comparing accuracy and speed with current popular methods, and an ablation study of the model construction process.

\begin{figure*}[ht]
\centering
\includegraphics[width=6.0in]{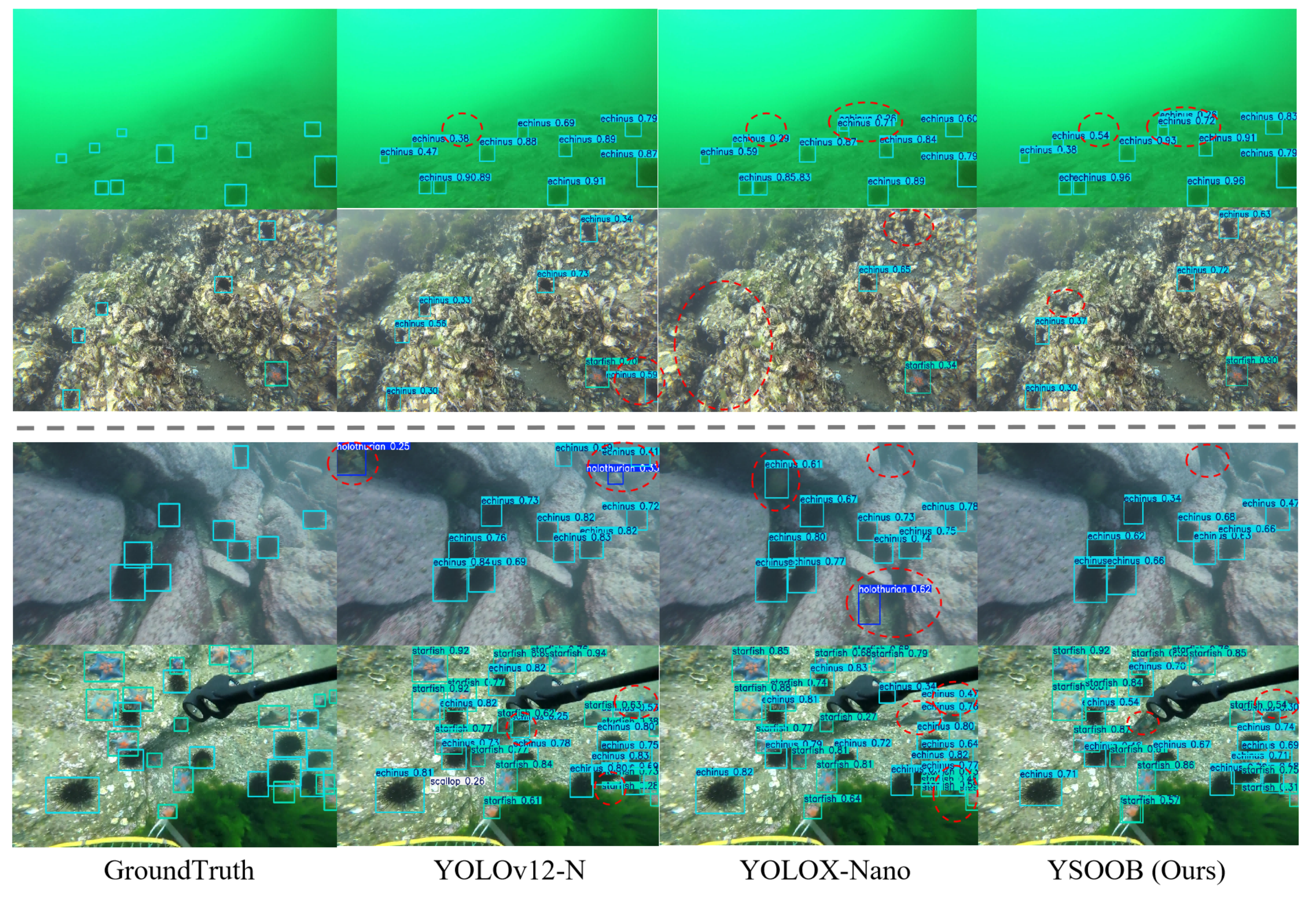}
\caption{Comparison of underwater object detection performance between the baseline model YOLOV12-N, our method YSOOB, and the parameter-similar YOLOX-Nano. The results above the dashed line are from the URPC2020 dataset, and those below are from the DUO dataset. The red dashed line indicates missed or incorrectly detected targets.}
\label{fig_1}
\end{figure*}

\subsection{Experimental Setup}
\subsubsection{Datasets}
We validated the proposed method on the URPC2020 dataset \cite{urpc2020}, which contains a total of 7,543 images across four different categories: holothurian, echinus, starfish, and scallop. After removing invalid images, the dataset was divided into 5,907 training samples and 1,476 validation samples. We also evaluated the model's generalization capability on the DUO dataset \cite{duo}, which consists of 6,671 training images and 1,111 test images, covering the same four categories of URPC2020.

\subsubsection{Implementation Details}
The method was implemented on a system equipped with an Intel(R) Xeon(R) Gold 6330 2.00GHz CPU, Ubuntu 20.04, and an NVIDIA RTX 4090 GPU, utilizing CUDA 12.4 and the PyTorch framework. In the experiments, training parameters were standardized across different models. The momentum was set to 0.937, the momentum decay coefficient was set to 0.0005, and the initial learning rate was set to 0.01. All models were trained on input images of size $640\times640$ for 200 epochs without pre-loaded weights. The latency for all models was tested using TensorRT FP16 on a T4 GPU \cite{v12} and Jetson Xavier NX, with a batch size set to 1.

\subsection{Comparison with State-of-the-arts}

In this section, we compare YSOOB with currently popular SOTA detectors on the URPC2020 and DUO datasets to verify the effectiveness of our method.

Table I presents the comparison results on the URPC2020 dataset. YSOOB achieves a mAP of 48.0\% with only 1.2M parameters, reaching a level comparable to that of YOLOs SOTA detectors with less than half the number of parameters. YOLOX-Nano \cite{yolox}, which also has ultra-light characteristics, suffers from an imbalance between efficiency and accuracy, failing to meet the accuracy requirements for practical applications. Due to its inefficient model framework, it also does not achieve the desired inference speed. In addition to inference tests on the T4 GPU, we deployed the model on the Jetson Xavier NX edge computing device to explore the practicality of different methods. YSOOB outperforms all other compared models with an inference latency of 17.3ms, achieving an excellent balance between accuracy and lightweight design, making it particularly suitable for real-time applications. Given the complexity of Faster-RCNN \cite{faster}, we did not conduct deployment tests for it. Although detectors based on the DETR framework \cite{rtdetr}, with global attention mechanisms, have achieved outstanding performance in various application scenarios, their FLOPs under the ResNet-R18 minimal framework still reach 60G, making them unsuitable for deployment on minimal hardware platforms. YSOOB, on the other hand, is 3× faster than DETR \cite{rtdetr}.

Fig. 3 visualizes the test results on the URPC2020 dataset. YOLOX-Nano's \cite{yolox} inefficient and overly minimal model parameters fail to fit the training data, leading to significant missed detections. In contrast, YSOOB performs comparably to the baseline model YOLOV12-N \cite{v12}, showing only a few minor errors.

\begin{table}[]
\centering
\Large
\caption{COMPARISON WITH OTHER DETECTORS ON DUO TEST SET}
\label{tab:my-table}
\renewcommand{\arraystretch}{1.2}
\resizebox{\columnwidth}{!}{%
\begin{tabular}{ccccccc}
\hline
\multirow{2}{*}{\textbf{Model}} &
  \multirow{2}{*}{\textbf{\begin{tabular}[c]{@{}c@{}}mAP50 \\ (\%)\end{tabular}}} &
  \multirow{2}{*}{\textbf{\begin{tabular}[c]{@{}c@{}}mAP\\ (\%)\end{tabular}}} &
  \multicolumn{4}{c}{\textbf{mAP (\%)}} \\ \cline{4-7} 
                                         &      &      & \textbf{Holothurian} & \textbf{Echinus} & \textbf{Starfish} & \textbf{Scallop} \\ \hline
YOLOv12-N \cite{v12} (Baseline)   & 83.1                      & 63.6                    & 59.5  & 74.3 & 73.1 & 47.5 \\
YOLOv8-N \cite{v8v11}     & 83.2 & \underline{63.8} & \underline{60.4}        & 74.0    & 73.2    & 47.6     \\
YOLOv10-N \cite{v10}    & \underline{83.3} & 63.6 & 59.5        & \underline{75.0}    & \underline{73.5}    & 46.4     \\
YOLOv11-N \cite{v8v11} & 83.0 & 63.4 & 60.3        & 74.1    & 72.5    & 46.7     \\
YOLOX-Nano  \cite{yolox}  & 52.6   & 39.7   & 39.8          & 43.6      & 41.8      & 33.6       \\
RT-DETRv2-R18 \cite{rtdetrv2} & \textbf{85.9}   & \textbf{66.8}   & \textbf{63.4}          & \textbf{78.1}      & \textbf{76.4}      & \textbf{49.3}       \\
\rowcolor{gray!20} \textbf{YSOOB (Ours)}      & 82.9   & 63.4   & 59.2          & 74.1      & 72.6      & \underline{47.7}       \\ \hline
\end{tabular}%
}
\end{table}

To further demonstrate YSOOB's generalization capability, we conducted experiments on the DUO dataset, as shown in Table II. The results indicate that YSOOB maintains performance slightly below that of SOTA detectors. Due to its refined characteristics, YSOOB performs exceptionally well in complex underwater multi-target scenarios. It rarely exhibits false detections and sometimes even outperforms YOLOV12-N \cite{v12}, as shown in Fig. 2.

\subsection{Ablation Experiment}

\begin{table}[]
\centering
\caption{ABLATION ON THE PROPOSED YSOOB}
\label{tab:my-table}
\renewcommand{\arraystretch}{1.2}
\resizebox{\columnwidth}{!}{%
\begin{tabular}{c|cccc|>{\columncolor{gray!15}}c}
\hline
\textbf{Ablations}   & \textbf{Baseline}  & \textbf{Ab1}  & \textbf{Ab2}  & \textbf{Ab3}  & \textbf{YSOOB}         \\ \hline
\textbf{WSME}        & -   & \checkmark    & -    & -    & \checkmark             \\
\textbf{Tran-Up}     & -    & -    & \checkmark    & -    & \checkmark             \\
\textbf{Even-Down}   & -    & -   & \checkmark    & -    & \checkmark             \\
\textbf{Chan-Comp}   & -    & -    & -    & \checkmark    & \checkmark             \\
\textbf{LKC-Recon}   & -    & -    & -    & \checkmark    & \checkmark             \\ \hline
\textbf{mAP50 (\%)}   & 83.2 & \textbf{83.7} & 83.1 & 82.7 & 83.1 \textcolor{red}{$\downarrow$ 0.12\%} \\
\textbf{mAP (\%)}     & 48.2 & \textbf{48.6} & 48.1 & 47.8 & 48.0 \textcolor{red}{$\downarrow$ 0.63\%}  \\
\textbf{\#Param. (M)} & 2.6  & 2.4  & 2.1  & 1.5  & \textbf{1.2} \textcolor[rgb]{0,0.7,0}{$\downarrow$ 53.85\%}  \\
\textbf{FLOPs (G)}   & 6.3  & 6.3  & 5.9  & 5.5  & \textbf{4.7} \textcolor[rgb]{0,0.7,0}{$\downarrow$ 25.40\%} \\ \hline
\end{tabular}%
}
\end{table}

We present the construction process of YSOOB in Table III. When we introduced WSME to the baseline model, the mAP improved by 0.83\%. WSME expands image features to the frequency domain, allowing the model to perceive spatial and frequency-domain variations simultaneously. It weakens the high-frequency components adaptively, which serves a similar function to UIE. When adjusting the model's resampling, even-sized convolution provided a significant advantage regarding model parameters. By pairing it with transposed convolution for dynamic spatial information modelling, the model achieved preliminary lightweight with almost no change in accuracy. In ablation 3, the model's parameters were reduced from 2.6M to 1.5M, and the accuracy slightly decreased. However, experience tells us that such trade-offs are acceptable in lightweight model design. Ultimately, the final model, YSOOB, achieved an impressive reduction of 53.85\% in the number of parameters and 25.40\% in FLOPs compared to the baseline model, with a minimal accuracy loss of less than 0.7\%.

\section{CONCLUSIONS}

This study proposes an Ultra-Light Real-Time Underwater Object Detection method, named YSOOB. To serve as an effective alternative to UIE, we employ a MSWE that directly extends the feature perception of the original degraded underwater image from the spatial domain to the frequency domain for encoding and reconstruction while performing high-frequency noise feature interaction and suppression. Additionally, we introduce even-sized convolution and transposed convolution to optimize the traditional resampling method in YOLOs, giving the model dynamic perception capabilities. Finally, we eliminate model redundancy through simple yet effective channel compression and RLKC, achieving a lightweight design. With only 1.2 millions parameters, YSOOB performs excellently in underwater object detection tasks, balancing performance and efficiency. Future research will optimize the model architecture and explore its generalization ability in broader real-world scenarios.

\addtolength{\textheight}{-12cm}  


\bibliographystyle{ieeetr}
\bibliography{ref}

\begin{thebibliography}{10}

\bibitem{stereo}
C.-W. Liu, Y.~Zhang, Q.~Chen, I.~Pitas, and R.~Fan, ``These maps are made by propagation: Adapting deep stereo networks to road scenarios with decisive disparity diffusion,'' {\em IEEE Transactions on Image Processing}, vol.~34, pp.~1516--1528, 2025.

\bibitem{ViPOcc}
Y.~Feng, Y.~Han, X.~Zhang, T.~Li, Y.~Zhang, and R.~Fan, ``Vipocc: Leveraging visual priors from vision foundation models for single-view 3d occupancy prediction,'' {\em Proceedings of the AAAI Conference on Artificial Intelligence}, vol.~39, pp.~3004--3012, Apr. 2025.

\bibitem{mfmos}
J.~Cheng, K.~Zeng, Z.~Huang, X.~Tang, J.~Wu, C.~Zhang, X.~Chen, and R.~Fan, ``Mf-mos: A motion-focused model for moving object segmentation,'' in {\em 2024 IEEE International Conference on Robotics and Automation (ICRA)}, pp.~12499--12505, IEEE, 2024.

\bibitem{1}
D.~Bogdoll, M.~Nitsche, and J.~M. Zollner, ``Anomaly detection in autonomous driving: A survey,'' in {\em 2022 IEEE/CVF Conference on Computer Vision and Pattern Recognition Workshops (CVPRW)}, p.~4487–4498, IEEE, June 2022.

\bibitem{leps}
X.~Huang, J.~Cheng, Q.~Xiang, J.~Dong, J.~Wu, R.~Fan, and X.~Tang, ``Leps: A lightweight and effective single-stage detector for pothole segmentation,'' {\em IEEE Sensors Journal}, vol.~24, no.~14, pp.~22045--22055, 2024.

\bibitem{3}
S.~Xu, M.~Zhang, W.~Song, H.~Mei, Q.~He, and A.~Liotta, ``A systematic review and analysis of deep learning-based underwater object detection,'' {\em Neurocomputing}, vol.~527, pp.~204--232, 2023.

\bibitem{edgeant}
J.~Dong, J.~Cheng, J.~Wu, C.~Zhang, S.~Zhao, and X.~Tang, ``Real-time aiot for uav antenna interference detection via edge-cloud collaboration,'' {\em IEEE Internet of Things Journal}, pp.~1--1, 2024.

\bibitem{5}
Y.~Wang, J.~Guo, W.~He, H.~Gao, H.~Yue, Z.~Zhang, and C.~Li, ``Is underwater image enhancement all object detectors need?,'' {\em IEEE Journal of Oceanic Engineering}, vol.~49, no.~2, pp.~606--621, 2024.

\bibitem{6}
J.~Wen, J.~Cui, G.~Yang, B.~Zhao, Z.~Gao, and B.~M. Chen, ``Underwater object detection integrated with image enhancement,'' in {\em 2024 IEEE International Conference on Real-time Computing and Robotics (RCAR)}, pp.~160--165, 2024.

\bibitem{7}
J.~Wen, J.~Cui, B.~Zhao, B.~Han, X.~Liu, Z.~Gao, and B.~M. Chen, ``Enyolo: A real-time framework for domain-adaptive underwater object detection with image enhancement,'' in {\em 2024 IEEE International Conference on Robotics and Automation (ICRA)}, pp.~12613--12619, 2024.

\bibitem{8}
K.~Mai, W.~Cheng, J.~Wang, S.~Liu, Y.~Wang, Z.~Yi, and X.~Wu, ``Underwater object detection based on dn-detr,'' in {\em 2023 IEEE International Conference on Real-time Computing and Robotics (RCAR)}, pp.~762--767, 2023.

\bibitem{9}
G.~Chen, Z.~Mao, Q.~Tu, and J.~Shen, ``A cooperative training framework for underwater object detection on a clearer view,'' {\em IEEE Transactions on Geoscience and Remote Sensing}, vol.~62, pp.~1--17, 2024.

\bibitem{10}
W.~Li, Y.~Li, R.~Li, H.~Shen, W.~Li, and K.~Yue, ``Research on rapid detection of underwater targets based on global differential model compression,'' {\em Journal of Marine Science and Engineering}, 2024.

\bibitem{12}
W.~Ouyang, Y.~Wei, and G.~Liu, ``A lightweight object detector with deformable upsampling for marine organism detection,'' {\em IEEE Transactions on Instrumentation and Measurement}, vol.~73, pp.~1--9, 2024.

\bibitem{13}
S.~Cheng, Y.~Han, Z.~Wang, S.~Liu, B.~Yang, and J.~Li, ``An underwater object recognition system based on improved yolov11,'' {\em Electronics}, vol.~14, no.~1, 2025.

\bibitem{14}
Y.~Zhong, B.~Li, L.~Tang, S.~Kuang, S.~Wu, and S.~Ding, ``Detecting camouflaged object in frequency domain,'' in {\em 2022 IEEE/CVF Conference on Computer Vision and Pattern Recognition (CVPR)}, pp.~4494--4503, 2022.

\bibitem{dcpi}
M.~Zhang, Y.~Feng, Q.~Chen, and R.~Fan, ``Dcpi-depth: Explicitly infusing dense correspondence prior to unsupervised monocular depth estimation,'' 2025.

\bibitem{15}
G.~Xu, W.~Liao, X.~Zhang, C.~Li, X.~He, and X.~Wu, ``Haar wavelet downsampling: A simple but effective downsampling module for semantic segmentation,'' {\em Pattern Recognition}, vol.~143, p.~109819, 2023.

\bibitem{16}
Z.~Zhang and D.~Zhang, ``Research on underwater object detection based on frequency domain attention mechanism,'' in {\em Fourth International Conference on Computer Vision, Application, and Algorithm (CVAA 2024)}, vol.~13486, pp.~327--334, SPIE, 2025.

\bibitem{17}
C.~Zhao, W.~Cai, C.~Dong, and C.~Hu, ``Wavelet-based fourier information interaction with frequency diffusion adjustment for underwater image restoration,'' in {\em 2024 IEEE/CVF Conference on Computer Vision and Pattern Recognition (CVPR)}, pp.~8281--8291, 2024.

\bibitem{v12}
Y.~Tian, Q.~Ye, and D.~Doermann, ``Yolov12: Attention-centric real-time object detectors,'' 2025.

\bibitem{19}
M.~Krichen, ``Generative adversarial networks,'' in {\em 2023 14th International Conference on Computing Communication and Networking Technologies (ICCCNT)}, pp.~1--7, 2023.

\bibitem{v8v11}
G.~Jocher, J.~Qiu, and A.~Chaurasia, ``{Ultralytics YOLO},'' Jan. 2023.

\bibitem{v10}
A.~Wang, H.~Chen, L.~Liu, K.~Chen, Z.~Lin, J.~Han, and G.~Ding, ``Yolov10: Real-time end-to-end object detection,'' {\em arXiv preprint arXiv:2405.14458}, 2024.

\bibitem{yolox}
Z.~Ge, S.~Liu, F.~Wang, Z.~Li, and J.~Sun, ``Yolox: Exceeding yolo series in 2021,'' {\em ArXiv}, vol.~abs/2107.08430, 2021.

\bibitem{rtdetrv2}
W.~Lv, Y.~Zhao, Q.~Chang, K.~Huang, G.~Wang, and Y.~Liu, ``Rt-detrv2: Improved baseline with bag-of-freebies for real-time detection transformer,'' {\em arXiv preprint arXiv:2407.17140}, 2024.

\bibitem{faster}
S.~Ren, K.~He, R.~Girshick, and J.~Sun, ``Faster r-cnn: Towards real-time object detection with region proposal networks,'' {\em IEEE Transactions on Pattern Analysis and Machine Intelligence}, vol.~39, no.~6, pp.~1137--1149, 2017.

\bibitem{mobilenetv3}
A.~Howard, M.~Sandler, G.~Chu, L.-C. Chen, B.~Chen, M.~Tan, W.~Wang, Y.~Zhu, R.~Pang, V.~Vasudevan, {\em et~al.}, ``Searching for mobilenetv3,'' in {\em Proceedings of the IEEE/CVF international conference on computer vision}, pp.~1314--1324, 2019.

\bibitem{channel}
Y.~Shen, L.~Shen, H.-Z. Huang, X.~Wang, and W.~Liu, ``Cpot: Channel pruning via optimal transport,'' {\em arXiv preprint arXiv:2005.10451}, 2020.

\bibitem{urpc2020}
C.~Liu, H.~Li, S.~Wang, M.~Zhu, D.~Wang, X.~Fan, and Z.~Wang, ``A dataset and benchmark of underwater object detection for robot picking,'' in {\em 2021 IEEE International Conference on Multimedia \& Expo Workshops (ICMEW)}, pp.~1--6, IEEE, 2021.

\bibitem{duo}
C.~{Liu}, H.~{Li}, S.~{Wang}, M.~{Zhu}, D.~{Wang}, X.~{Fan}, and Z.~{Wang}, ``{A Dataset And Benchmark Of Underwater Object Detection For Robot Picking},'' {\em arXiv e-prints}, p.~arXiv:2106.05681, June 2021.

\bibitem{rtdetr}
Y.~Zhao, W.~Lv, S.~Xu, J.~Wei, G.~Wang, Q.~Dang, Y.~Liu, and J.~Chen, ``Detrs beat yolos on real-time object detection,'' in {\em Proceedings of the IEEE/CVF conference on computer vision and pattern recognition}, pp.~16965--16974, 2024.

\end{thebibliography}

\end{document}